\documentclass[10pt,twocolumn,letterpaper]{article}

\usepackage{cvpr}
\usepackage{times}
\usepackage{epsfig}
\usepackage{graphicx}
\usepackage{amsmath}
\usepackage{amssymb}
\usepackage{diagbox}

\usepackage{graphicx}       
\usepackage{subfig}         
\usepackage{multirow}       
\usepackage{algorithm}      
\usepackage{algorithmic}    
\usepackage{lipsum}    

\usepackage[breaklinks=true,bookmarks=false]{hyperref}

\cvprfinalcopy 

\def\cvprPaperID{****} 

\begin{document}

\title{Quantization Networks}

%

\author{
Jiwei Yang\textsuperscript{1, 3}\thanks{This work was done when the author
was visiting Alibaba as a research intern.},
Xu Shen\textsuperscript{3},
Jun Xing\textsuperscript{2},
Xinmei Tian\textsuperscript{1}\thanks{Corresponding author.},
Houqiang Li\textsuperscript{1},\\
Bing Deng\textsuperscript{3},
Jianqiang Huang\textsuperscript{3},
Xiansheng Hua\textsuperscript{3}\thanks{Corresponding author.}\\
\textsuperscript{1}University of Science and Technology of China\\
\textsuperscript{2}University of Southern California\\
\textsuperscript{3}Alibaba Group\\
{\tt\small yjiwei@mail.ustc.edu.cn, junxnui@gmail.com, \{xinmei,lihq\}@ustc.edu.cn,}\\
{\tt\small \{shenxu.sx,dengbing.db,jianqiang.hjq,xiansheng.hxs\}@alibaba-inc.com}
}

\maketitle

\begin{abstract}
Although deep neural networks are highly effective, their high computational
and memory costs severely challenge their applications on portable devices. As
a consequence, low-bit quantization, which converts a full-precision neural
network into a low-bitwidth integer version, has been an active and promising
research topic. Existing methods formulate the low-bit quantization of networks
as an approximation or optimization problem. Approximation-based methods
confront the gradient mismatch problem, while optimization-based methods are
only suitable for quantizing weights and could introduce high computational
cost in the training stage. In this paper, we propose a novel perspective of
interpreting and implementing neural network quantization by formulating
low-bit quantization as a differentiable non-linear function (termed
quantization function). The proposed quantization function can be learned in a
lossless and end-to-end manner and works for any weights and activations of
neural networks in a simple and uniform way. Extensive experiments on image
classification and object detection tasks show that our quantization networks
outperform the state-of-the-art methods. We believe that the proposed method
will shed new insights on the interpretation of neural network quantization.
Our code is available at \url{https://github.com/aliyun/alibabacloud-quantization-networks}.

\end{abstract}


\section{Introduction}
\label{sec:introduction}
\begin{figure}[!h]
	\centering
	\subfloat[Sigmoid]{
		\includegraphics[width=0.4\linewidth]{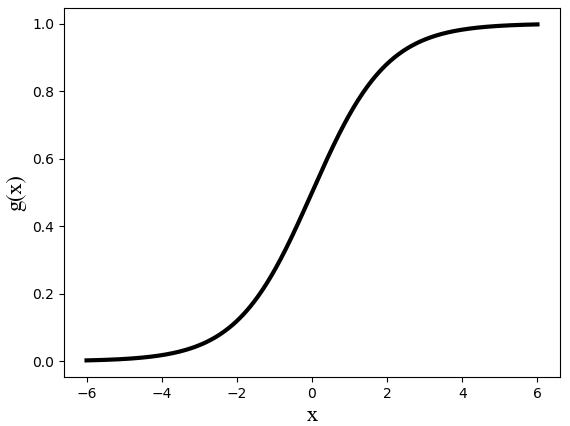}
		\label{fig:sigmoid}
	}
	\subfloat[ReLU]{
		\includegraphics[width=0.4\linewidth]{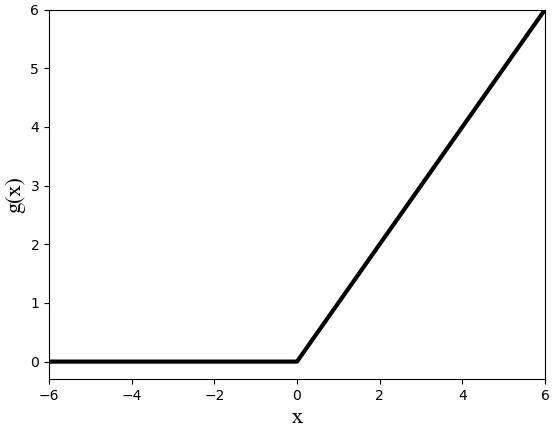}
		\label{fig:ReLU_function}
	}

	\subfloat[Maxout]{
		\includegraphics[width=0.4\linewidth]{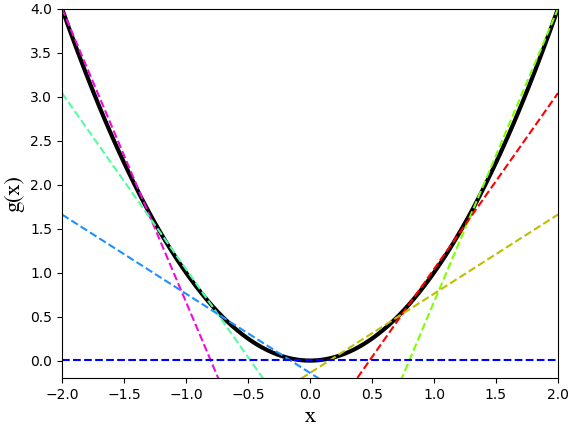}
		\label{fig:maxout}
	}
	\subfloat[Quantization]{
		\includegraphics[width=0.4\linewidth]{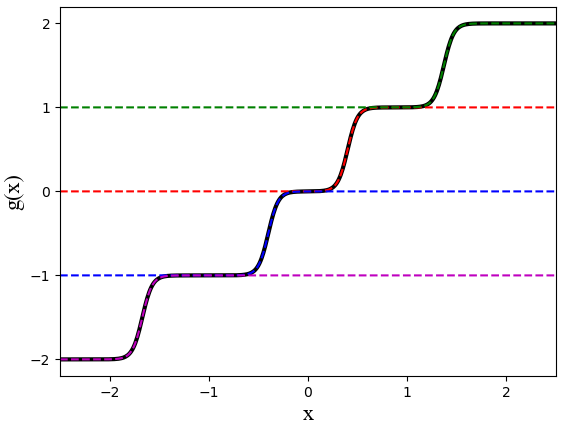}
		\label{fig:quan_function}
	}
	\caption{Non-linear functions used in neural networks.}
	\label{fig:functions}
\end{figure}
Although deep neural networks (DNNs) have achieved huge success in various domains,
their high computational and memory costs prohibit their deployment in scenarios where both computation and storage resources are limited. Thus, the democratization of deep learning hinges
on the advancement of efficient DNNs.
Various techniques have been proposed to lighten DNNs by either reducing the
number of weights and connections or by quantizing the weights and activations to lower bits.
As exemplified by ResNet \cite{ResNet}, SqueezeNet \cite{SqueezeNet} and MobileNet
\cite{MobileNet}, numerous efforts have been devoted to designing networks with compact
layers and architectures. Once trained, these networks can be further compressed
with techniques such as network pruning \cite{Han:2015:LBW},
weight sharing \cite{HashedNets} or matrix factorization \cite{JaderbergVZ14}.

Approaches for quantizing full-precision networks into low-bit
networks can be roughly divided into two categories: approximation-based and optimization-based.
Methods in the first category approximate the full-precision (32-bit) values with discrete low-bit (e.g. binary)
values via step functions in the forward pass \cite{xnor-net,integer-net,DoReFa,ternary-residual,TTQ,integer-arithmetic,twn,HORQ,HWGQ}.
Because the gradients of such approximations are saturated,  additional approximations in the backward process are needed.
As a consequence, the use of different forward and backward approximations causes a
gradient mismatch problem, which makes the optimization unstable. To avoid
the approximation of gradients, some methods
formulate the quantization of neural networks as a discretely constrained optimization problem,
where losses of the networks are incorporated \cite{admm,loss-aware}. Unfortunately,
optimization-based methods are only suitable for the quantization of weights.
Moreover, the iterative solution of the optimization problem suffers from a high
computational complexity during training.

Intuitively, if we can formulate the quantization operation as a simple non-linear function
similar to the common activation functions (e.g., Sigmoid \cite{sigmoid}, ReLU
\cite{relu} or Maxout \cite{maxout}), no approximation of gradients would be needed, and the
quantization of any learnable parameters in DNNs, including activations and weights, can be
learned straightforwardly and efficiently.
Inspired by that, we present a novel perspective for interpreting and implementing quantization in neural networks.
Specifically, we formulate quantization as a
differentiable non-linear mapping function, termed quantization function.
As shown in Fig.~\ref{fig:functions}, the quantization function is formed as a linear combination of several Sigmoid functions
with learnable biases and scales. In this way, the proposed quantization function can be learned in a lossless and end-to-end
manner and works for any weights and activations in neural networks, avoiding the gradient mismatch problem.
As illustrated in Fig.~\ref{fig:process}, the quantization is achieved via the continuous relaxation of the steepness of the Sigmoid functions during the training stage.

Our main contributions are summarized as follows:
\begin{itemize}
	\item In contrast to existing low-bit quantization methods,
	we are the first to formulate quantization as a differentiable non-linear mapping function, which provides a \emph{simple/straightforward} and \emph{general/uniform} solution for any-bit weight and activation quantization, without suffering the severe gradient mismatch problem.
	\item We implement a simple and effective form of quantization networks, which could be learned in a lossless and end-to-end manner and outperform state-of-the-art quantization methods on both image classification and object detection tasks.
\end{itemize}

\section{Related Work}
\label{sec:related_works}
In this paper, we propose formulating the quantization operation as a
differentiable non-linear function. In this section, we give a brief review of
both low-bit quantization methods and non-linear functions used in neural
networks.

\subsection{Low-Bit Quantization of Neural Networks}
Approaches for quantizing full-precision networks into low-bit
networks can be roughly divided into two categories: approximation-based and optimization-based.
The first approach is to approximate the 32-bit full-precision values with discrete low-bit
values in the forward pass of networks.
BinaryConnect \cite{BinaryConnect}
directly optimizes the loss of the network with weights $W$ replaced by
sign($W$), and approximates the sign function with the ``hard tanh'' function in the backward process,
to avoid the zero-gradient problem. Binary weight network (BWN) \cite{xnor-net} adds scale factors for
the weights during binarization. Ternary weight
network (TWN) \cite{twn} introduces ternary weights and
achieves better performance. Trained ternary quantization (TTQ)
\cite{TTQ} proposes learning both ternary values and scaled
gradients for $32$-bit weights. DoReFa-Net \cite{DoReFa} proposes quantizing
$32$-bit weights, activations and gradients using different widths of bits.
Gradients are approximated by a custom-defined form based on the mean of the absolute
values of full-precision weights. In \cite{integer-net}, weights, activations,
gradients and errors are all approximated by low-bitwidth integers
based on rounding and shifting operations. Jacob et al.
\cite{integer-arithmetic} propose an affine mapping of integers to real
numbers that allows inference to be performed using integer-only arithmetic.
As discussed before, the approximation-based methods use different forward and backward
approximations, which causes a gradient mismatch problem.
Friesen and Domingos \cite{iclr18} observe that setting targets
for hard-threshold hidden units to minimize loss is a discrete optimization problem.
Zhuang et al. \cite{telcnn} propose a two-stage approach to quantize the weights and activations in a two-step manner.
Lin et al. \cite{abcnet} approximate full-precision weights with the linear combination of multiple binary weight bases.
Zhang et al. \cite{lqnet} propose an flexible un-uniform quantization method to quantize both network weights and activations.
Cai et al. \cite{HWGQ} used several piece-wise backward approximators to overcome the problem of gradient mismatch.
Zhou et al. \cite{INQ} proposed a decoupling step-by-step operation to efficiently convert a pre-trained
full-precision convolutional neural network (CNN) model into a low-precision version.
As a specific quantization, HashNet \cite{HashNet} adopts a similar continuous relaxation to train the hash function, where a
single tanh function is used for binarization. However, our training case
(multi-bits quantization of both activations and weights in multi-layers) is
much more complicated and challenging.

To avoid the gradient approximation problem, optimization-based quantization methods are recently proposed.
They directly formulate the quantization of neural networks as a discretely constrained
optimization problem \cite{admm,loss-aware}. Leng et al.
\cite{admm} introduce convex linear constraints for the weights
and solve the problem by the alternating direction method of multipliers (ADMM).
Hou and Kwok \cite{loss-aware} directly optimize the loss function w.r.t. the
ternarized weights using proximal Newton algorithm.  However, these methods are only suitable for
quantization of weights and such iterative solution suffers from high computational costs in training.

\begin{figure*}[!tb]
	\centering
	\subfloat[No Quantization]{
		\includegraphics[width=0.19\linewidth]{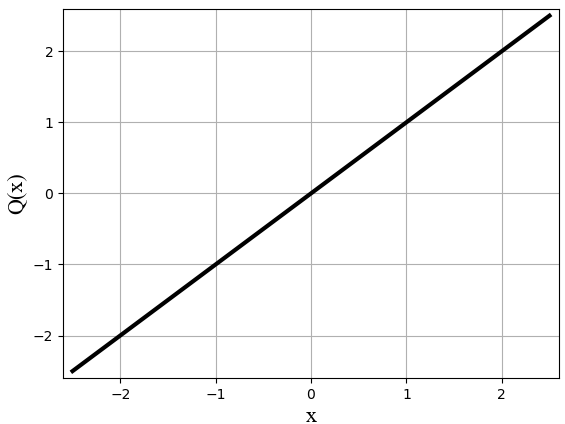}
		\label{fig:no-q}
	}
	\subfloat[T=1]{
		\includegraphics[width=0.19\linewidth]{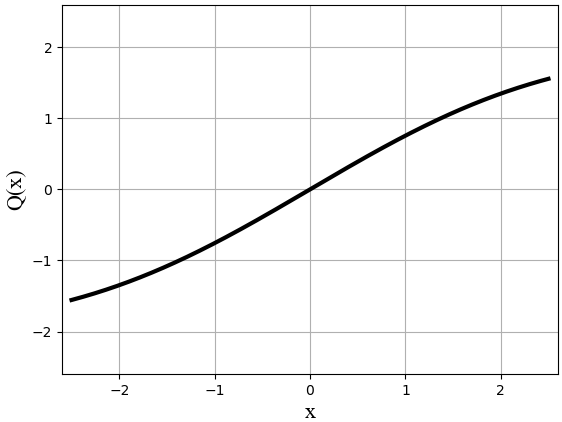}
		\label{fig:t-1}
	}
	\subfloat[T=11]{
		\includegraphics[width=0.19\linewidth]{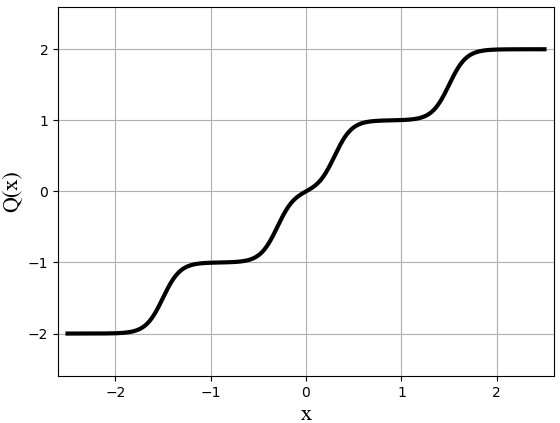}
		\label{fig:t-11}
	}
	\subfloat[T=121]{
		\includegraphics[width=0.19\linewidth]{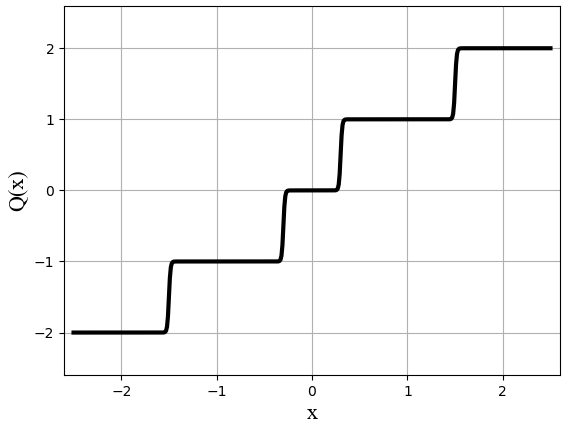}
		\label{fig:t-121}
	}
	\subfloat[Complete Quantization]{
		\includegraphics[width=0.19\linewidth]{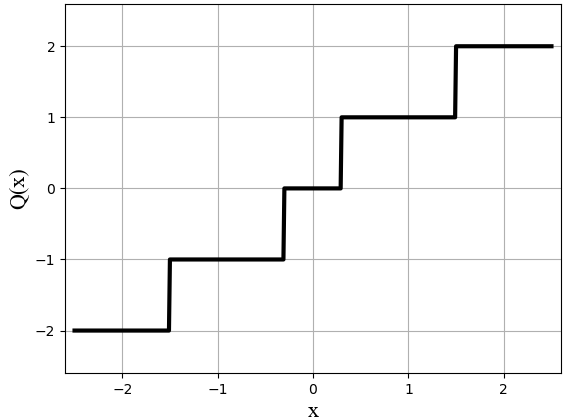}
		\label{fig:t-591}
	}
	\caption{The relaxation process of a quantization function during training, which goes from a straight line to steps as the temperature T increases.}
	\label{fig:process}
\end{figure*}

\subsection{Non-Linear Functions in Deep Neural Networks}
In neural networks, the design of hidden units is distinguished by the choice of the non-linear activation function $g(x)$ for hidden units
\cite{deep-learning-book}.
The simplest form of a neural network is perceptron \cite{perceptron}, where
a unit step function is introduced to produce a binary output:
\begin{equation}
g(x) = \mathcal{A}(x) = \left \{\begin{array}{cc}
1 & x \geq 0, \\
0 & x < 0.
\end{array} \right.
\end{equation}
This form is similar to the binary quantization operation, i.e., discretize the
continuous inputs into binary values. However, the problem is that it is not
immediately obvious how to learn the perceptron networks \cite{NNDL-book}.

To solve this problem, the
sigmoid activation function is adopted in the early form of feedforward neural networks:
\begin{equation}
g(x) = \sigma(x) = \frac{1}{1 + exp(-x)},
\end{equation}
which has smooth and non-zero gradient everywhere so that the sigmoid neurons can be learned via back-propagation.
When the absolute value of $x$ is very large, the outputs of a sigmoid function is close to a unit step function.

Currently, rectified linear units (ReLU) are more frequently used as the activation functions in deep
neural networks:
\begin{equation}
g(x) = ReLU(x) = \max(0, x).
\end{equation}
The ReLU function outputs zero across half of its domain and is linear in the other half, which makes the DNNs easy to optimize.

A generalization of the rectified linear units is Maxout. Its activation function is defined as:
\begin{equation}
g(x) = \underset{j}\max ~(a_j*x + c_{j}), j=1, \dots, k
\end{equation}
where \{$a_j$\} and \{$c_j$\} are learned parameters. The form of Maxout indicates that a complex convex function can be
approximated by a combination of $k$ simple linear functions.

\section{Quantization Networks}
\label{sec:quantization-networks}
The main idea of this work is to formulate the quantization operation as a differentiable
non-linear function, which can be applied to any weights and activations in
deep neural networks.
We first present our novel interpretation of quantization from the perspective of non-linear functions.
Then, our simple and effective quantization function is introduced and the learning of
quantization networks are given.

\subsection{Reformulation of Quantization}
The quantization operation is mapping continuous inputs into discrete
integer numbers, which is similar to the perceptron. Thus, from the perspective of non-linear mapping
functions, a binary quantization operation can be formed of a unit step function. Inspired by
the design of Maxout units, quantizing continuous values into a set of integer
numbers can be formulated as a combination of several binary quantizations. In other words,
the ideal low-bit quantization function is a combination of several unit step functions with
specified biases and scales, as shown in Fig. \ref{fig:process}(e):
\begin{equation}
\label{eq:step-functions}
y = \sum_{i=1}^n s_i \mathcal{A}(\beta x - b_i) - o,
\end{equation}
where $x$ is the full-precision weight/activation to be quantized, $y$ is the quantized
integer constrained to a predefined set $\mathcal{Y}$, and $n+1$ is the number of
quantization intervals. $\beta$ is the scale factor of inputs.
$\mathcal{A}$ is the standard unit step function. $s_i$ and $b_i$ are the scales and biases for
the unit step functions, $s_i = \mathcal{Y}_{i+1} - \mathcal{Y}_i$. The global offset $o = \frac{1}{2}\sum_{i=1}^n s_i$ keeps the quantized output zero-centered.
Once the expected quantized integer set $\mathcal{Y}$ is given, $n=|\mathcal{Y}|-1$, $s_i$ and offset $o$ can be directly obtained.

For example, for a 3-bit quantization, the output $y$ is restricted to $\mathcal{Y}=\{-4, -2, -1, 0, 1, 2,
4\}$, $n=|\mathcal{Y}|-1=6$, $\{s_i\} = \{2, 1, 1, 1, 1, 2\}$, and $o=4$. $\beta$ and $b_i$ are parameters to be learned.
Because the step function is not smooth, it is not immediately
obvious how we can learn a feedforward networks with Eq. (\ref{eq:step-functions}) applied to
activations or weights \cite{NNDL-book}.

\subsection{Training and Inference with Quantization Networks}
Inspired by the advantage of sigmoid units against the perceptron in feedforward
networks, we propose replacing the unit step functions in the ideal quantization function
Eq. (\ref{eq:step-functions}) with sigmoid functions. With this replacement, we can have a
differentiable quantization function, termed soft quantization function, as shown in Fig. \ref{fig:process}(c). Thus, we can learn
any low-bit quantized neural networks in an end-to-end manner based on back propagation.

However, the ideal quantization function Eq. (\ref{eq:step-functions}) is applied in the inference
stage. The use of different quantization functions in training and inference stages may decrease the
performance of DNNs. To narrow the gap between the ideal quantization function used in inference stage
and the soft quantization function used in training stage,
we introduce a temperature $T$ to the sigmoid function, motivated by the temperature
introduced in distilling \cite{distilling},
\begin{equation}
\sigma(Tx) = \frac{1}{1 + exp(-Tx)}.
\end{equation}
With a larger $T$, the gap between two quantization functions is smaller, but the learning capacity
of the quantization networks is lower since the gradients of the soft quantization function will be
zero in more cases. To solve this problem, in the training stage we start with a small $T$ to ensure
the quantized networks can be well learned, and then gradually increase $T$ w.r.t. the training epochs.
In this way, the quantized networks can be well learned and the gap between two quantization functions
will be very small at the end of the training.


\textbf{Forward Propagation}.
In detail, for a set of full-precision weights or activations to be quantized
$\mathcal{X}=\{x_d, d = 1, \cdots, D\}$, the quantization
function is applied to each $x_d$ independently:
\begin{equation}
\label{eq:quantization}
y_d = \mathcal{Q}(x_d) = \alpha (\sum_{i=1}^n s_i \sigma(T(\beta x_d - b_i)) - o),
\end{equation}
where $\beta$ and $\alpha$ are the scale factors of the input and output respectively.
$\mathbf{b} = [b_i, i = 1, \cdots, n]$, where $b_i$ indicates the beginning of
the input for the $i$-th quantization interval except the first quantization interval, and the beginning of the first quantization interval is $-\infty$. The temperature $T$ controls the gap between the ideal
quantization function and the soft quantization function.
The gradual change from no quantization to complete quantization along with the adjustment of $T$ is
depicted in Fig. \ref{fig:process}.

The quantization function Eq. (\ref{eq:quantization}) is applied to every full-precision value
$x$ that need to be quantized, just as applying ReLU in traditional DNNs. $x$ can be either a
weight or an activation in DNNs. The output $y$ replaces $x$ for further computing.


\textbf{Backward Propagation}.
During training stage, we need to back propagate the
gradients of loss $\ell$ through the quantization function, as well as compute the
gradients with respect to the involved parameters :

%
%
%

\begin{align}
\frac{\partial \ell}{\partial x_d} & = \frac{\partial \ell}{\partial
	y_d} \cdot
\sum_{i=1}^n \frac{T\beta}{\alpha s_i} g_d^i (\alpha s_i - g_d^i), \\
\frac{\partial \ell}{\partial \alpha} & = \sum_{d=1}^D
\frac{\partial \ell}{\partial y_d} \cdot \frac{1}{\alpha} y_d, \\
\frac{\partial \ell}{\partial \beta} & = \sum_{d=1}^D
\frac{\partial \ell}{\partial y_d} \cdot \sum_{i=1}^n
\frac{T x_d}{\alpha s_i} g_d^i (\alpha s_i - g_d^i), \\
\frac{\partial \ell}{\partial b_i} & = \sum_{d=1}^D
\frac{\partial \ell}{\partial y_d} \cdot
\frac{-T}{\alpha s_i} g_d^i(\alpha s_i - g_d^i).
\end{align}
where $g_d^i = \sigma(T(\beta x_d - b_i))$.
we do not need to compute the gradients of  $n$, $s_i$ and offset $o$,
because their are directly obtained by $\mathcal{Y}$.
Our soft quantization function is a differentiable transformation that introduces quantized weights and
activations into the network.

\begin{algorithm}
	\caption{Training quantization networks}
	\label{alg:training}
	\begin{algorithmic}
		\STATE \textbf{Input:} Network $N$ with $M$ modules $\mathcal{M}_{m=1}^M$
		and their corresponding activations/inputs $\{\mathcal{X}^{(m)}\}_{m=1}^M$ and trainable weights (or other parameters) $\{\Theta^{(m)}\}_{m=1}^M$
		\STATE \textbf{Output:} Quantized network for inference, $N_{Q}^{inf}$
		\STATE $N_Q^{tr} \gets N$ // Training quantization network
		\FOR{$epoch \gets 1 \textrm{ to } Max\_Epochs$}
		\FOR{$m \gets 1 \textrm{ to } M$}
		\STATE Apply the soft quantization function to each element $x^{m}_d$ in $\mathcal{X}^{(m)}$ and each element $\theta^{m}_d$ in $\Theta^{(m)}$: \\
		$y^{m}_d = \mathcal{Q}_{\{\alpha^{(m)}_\mathcal{X}, \beta^{(m)}_\mathcal{X},\beta^{(m)}_\Theta \} }(x^{m}_d)$, \\
		$\widehat{\theta}^{m}_d = \mathcal{Q}_{\{\alpha^{(m)}_\Theta, \beta^{(m)}_\Theta,
			b^{(m)}_\Theta \}}(\theta^{m}_d)$. \\
		Forward propagate module $m$ with the quantized weights and activations.
		\ENDFOR
		\ENDFOR
		\STATE Train $N_Q^{tr}$ to optimize the parameters
		$\Theta \cup {\{\alpha^{(m)}_\Theta,
			\beta^{(m)}_\Theta, b^{(m)}_\Theta, \alpha^{(m)}_\mathcal{X},
			\beta^{(m)}_\mathcal{X}, b^{(m)}_\mathcal{X}\}}_{m=1}^M$ with gradually increased temperature
		$T$
		\STATE $N_Q^{inf} \gets N_Q^{tr}$ // Inference quantization network with
		frozen parameters
		\FOR{$m \gets 1 \textrm{ to } M$}
		\STATE Replace the soft quantization functions with Eq. (\ref{eq:inference}) for inference.
		\ENDFOR
	\end{algorithmic}
\end{algorithm}

\begin{table*}[!tb]
	\begin{center}
		\begin{tabular}{|c|c|c|c|c|c|c|c|}
			\hline
			\diagbox{Methods}{W/A} & 1/32 & 2/32 & 3($\pm$2)/32 & 3($\pm$4)/32 & 1/1  & 1/2 \\
			\hline\hline
			BinaryConnect \cite{BinaryConnect} & 35.4/61.0 & - & - & - & 27.9/50.42  & - \\
			BWN \cite{xnor-net} & 56.8/79.4 & - & - & - & 44.2/69.2 & - \\
			DoReFa \cite{DoReFa} & 53.9/76.3 & - & - & - & 39.5/- & 47.7/- \\
			TWN \cite{twn} & - & 54.5/76.8 & - & - & - & -  \\
			TTQ \cite{TTQ} & - & 57.5/79.7 & - & - & - & - \\
			ADMM \cite{admm} & 57.0/79.7 & 58.2/80.6 & 59.2/81.8 & 60.0/82.2 & - & - \\
			HWGQ \cite{HWGQ} & - & - & - & - & - & 52.7/76.3 \\
			TBN \cite{TBN} & - & - & - & - & - & 49.7/74.2 \\
			LQ-Net \cite{lqnet} & - & 60.5/82.7 & - & - & - & \textbf{55.7/78.8} \\
			Ours & \textbf{58.8/81.7} & \textbf{60.9/83.2} & \textbf{61.5/83.5}
			& \textbf{61.9/83.6} & \textbf{47.9/72.5} & 55.4/\textbf{78.8} \\
			\hline
		\end{tabular}
	\end{center}
	\caption{Top-1 and Top-5 accuracies (\%) of AlexNet on ImageNet classification. Performance of
		the full-precision model is $\textbf{61.8/83.5}$. ``W" and ``A" represent the quantization bits of weights and activations, respectively.}
	\label{tab:alexnet}
\end{table*}

\textbf{Training and Inference}. To quantize a network, we specify a set of
weights or activations and insert the quantization function for each of
them, according to Eq. (\ref{eq:quantization}). Any layer that previously
received $x$ as an input, now receives $\mathcal{Q}(x)$. Any module that
previously used $W$ as parameters, now uses $\mathcal{Q}(W)$. The smooth
quantization function $\mathcal{Q}$ allows efficient training for networks,
but it is neither necessary nor desirable during inference; we want the specified
weights or activations to be discrete numbers. For this, once the network
has been trained, we replace the sigmoid function in Eq. (\ref{eq:quantization}) by the unit step
function for quantization:
\begin{equation}
\label{eq:inference}
y = \alpha (\sum_{i=1}^n s_i \mathcal{A}(\beta x - b_i) - o).
\end{equation}

Algorithm \ref{alg:training} summarizes the procedure for training quantization networks. For a full-precision
network $N$ with $M$ modules, where a module can be either a convolutional layer or a fully connected layer,
we denote all the activations to be quantized in the $m$-th module as $\mathcal{X}^{(m)}$, and denote all the
weights to be quantized in the $m$-th module as $\Theta^{(m)}$. All elements in $\mathcal{X}^{(m)}$ share the
same quantization function parameters $\{\alpha^{(m)}_\mathcal{X}, \beta^{(m)}_\mathcal{X},
\mathbf{b}^{(m)}_\mathcal{X}\}$. All elements in $\Theta^{(m)}$ share the same quantization function parameters
$\{\alpha^{(m)}_\Theta, \beta^{(m)}_\Theta, \mathbf{b}^{(m)}_\Theta\}$.
We apply the quantization function module by module. Then, we train the network with gradually increased temperature $T$.

\section{Experiments}
\label{sec:experiments}

\subsection{Image Classification}
\label{sec:classification}
To compare with state-of-the-art methods, we evaluate our method on ImageNet
(ILSVRC 2012). ImageNet has approximately $1.2$ million training images from $1$ thousand categories and
$50$ thousand validation images. We evaluate our method on AlexNet \cite{alexnet}
(over-parameterized architectures) and ResNet-18/ResNet-50 \cite{ResNet}
(compact-parameterized architectures). We report our classification performance
using Top-1 and Top-5 accuracies with networks quantized to Binary(\{0, 1\}, 1 bit), Ternary(\{-1, 0, 1\}, 2 bits), \{-2, -1, 0, 1, 2\}
(denoted as 3 bits($\pm$2)), \{-4, -2, -1, 0, 1, 2, 4 \} (denoted as 3 bits($\pm$4)), and \{-15, -14, $\cdots$, -1, 0, 1, $\cdots$, 14, 15 \} (5 bits). All the
parameters are fine-tuned from pretrained full-precision models.

All the images from ImageNet are resized to have $256$ pixels for the smaller
dimension, and then a random crop of $224\times224$ is selected for training.
Each pixel of the input images is subtracted by the mean values and divided
by variances. Random horizontal flipping is introduced for preprocessing.
No other data augmentation tricks are used in the learning process. The batch size
is set to $256$. Following \cite{xnor-net} and \cite{twn}, the parameters of the
first convolutional layer and the last fully connected layer for classification
are not quantized. For testing, images are resized to $256$ for the smaller side,
and a center crop of $224\times 224$ is selected.

For our quantization function Eq. (\ref{eq:quantization}),
to ensure all the input full-precision values lie in the linear region of our quantization
function, the input scale $\beta$ is initialized as $\frac{5p}{4} \times \frac{1}{q}$,
where $p$ is the max absolute value of elements in $\mathcal{Y}$ and $q$ is the max
absolute value of elements in $\mathcal{X}$. The output
scale $\alpha$ is initialized by $\frac{1}{\beta}$, keeping the magnitude of
the inputs unchanged after quantization.

\textbf{Weight quantization}: For binary quantization, only $1$
sigmoid function is needed; thus $n=1$, $b=0$, $s=2$, and $o=1$. For ternary quantization
($\{-1, 0, 1\}$), $n=2$, $s_i=1$ and $b_1=-0.05, b_2=0.05$, ensuring that $5\%$ of the values in
$[-1, 1]$ are quantized to $0$ as in \cite{twn}. For the quantization of other
bits, we first group the full-precision inputs into $n+1$ clusters by $k$-means
clustering. Then, the centers of the clusters are ranked in ascending order, and we get $\{c_1, \dots, c_{n+1}\}$.  For bias initialization, $b_i = \frac{c_i + c_{i+1}}{2}$. Again, we set
$s_{\lfloor\frac{n}{2}\rfloor} =
s_{\lfloor\frac{n}{2}\rfloor + 1} = 1$ and $b_{\lfloor\frac{n}{2}\rfloor}
= -0.05, b_{\lfloor\frac{n}{2}\rfloor+1} = 0.05$ to ensure that $5\%$ of the values in $[-1, 1]$ are quantized to $0$.

\begin{table*}[!tb]
	\begin{center}
		\begin{tabular}{|c|c|c|c|c|c|c|c|c|c|}
			\hline
			\diagbox{Methods}{W/A} & 1/32 & 2/32 & 3($\pm$2)/32 & 3($\pm$4)/32 &   5/32  & 1/1 & 1/2 & 32/2 \\
			\hline \hline
			BWN \cite{xnor-net} & 60.8/83.0 & - & - & - &  -  & 51.2/73.2 & - & -\\
			TWN \cite{twn} & - & 61.8/84.2 & - & - & - & - & - & -  \\
			TTQ \cite{TTQ} & - & 66.6/87.2 & - & - & -  & - & - & -  \\
			INQ \cite{INQ} & - & 66.0/87.1 & - & 68.1/88.4 & 69.0/89.1 & - & - & -  \\
			ABC-Net \cite{abcnet} & - & - & - & - & 68.3/87.9 & 42.7/67.6 & - & - \\
			HWGQ \cite{HWGQ} & - & - & - & - & - & - & 59.6/82.2 & -  \\
			ADMM \cite{admm} & 64.8/86.2 & 67.0/87.5 & 67.5/87.9 & 68.0/88.3 &
			- & - & - & - \\

			ICLR18 \cite{iclr18} & - & - & - & - & - & - &  - & 64.3/- \\

			TBN \cite{TBN} & - & - & - & - & - & - & 55.6/79.0 & - \\
			LQ-Net \cite{lqnet} & - & 68.0/88.0 & - & 69.3/88.8 & - & - & 62.6/84.3 & - \\
			Ours & \textbf{66.5/87.3} & \textbf{69.1/88.9} & \textbf{69.9/89.3}
			& \textbf{70.4/89.6} & \textbf{70.6/89.6}  & \textbf{53.6/75.3} &  \textbf{63.4/84.9} & \textbf{65.7/86.5} \\
			\hline
		\end{tabular}
	\end{center}
	\caption{Top-1 and Top-5 accuracies (\%) of ResNet-18 on ImageNet classification. Performance of
		the full-precision model are $\textbf{70.3/89.5}$. ``W" and ``A" represent the quantization bits of weights and activations, respectively.}
	\label{tab:resnet}
\end{table*}

\begin{table*}[!tb]
	\begin{center}
		\begin{tabular}{|c|c|c|c|c|c|c|}
			\hline
			\diagbox{Methods}{W/A} & 1/32 & 2/32 & 3($\pm$2)/32 & 3($\pm$4)/32 &   5/32   \\
			\hline \hline
			BWN \cite{xnor-net} & 68.7/- & - & -  & - & -  \\
			TWN \cite{twn} & - & 72.5/- & - & - & - \\
			INQ \cite{INQ} & - & - & - & - & 74.8/-  \\
			LQ-Net \cite{lqnet} & - & 75.1/92.3 & - & - & -   \\
			Ours & \textbf{72.8/91.3} & \textbf{75.2}/\textbf{92.6}
			& \textbf{75.5/92.8} & \textbf{76.2/93.2}  & \textbf{76.4/93.2} \\
			\hline
		\end{tabular}
	\end{center}
	\caption{Top-1 and Top-5 accuracies (\%) of ResNet-50 on ImageNet classification. Performance of
		the full-precision model are $\textbf{76.4/93.2}$. ``W" and ``A" represent the quantization bits of weights and activations, respectively.}
	\label{tab:resnet50}
\end{table*}

\textbf{Activation quantization}: Outputs of the ReLU units are used for activation quantization. It means that the block is Conv−-BN−-ReLU(-−Pooling)-−Quant in our method. The $o$ in Eq. (\ref{eq:quantization}) is set to $0$ because all activations are non-negative. For binary quantization($\{0, 1\}$), only $1$ sigmoid function is needed, \ie $n=1$ and $s=1$. For two-bit quantization of activations ($\{0, 1, 2, 3\}$), $n=3$ and $s_i=1$. $b_i$ is obtained by clustering as in weight quantization.
We randomly sample 1000 samples from the dataset, and get the min/max
activation values of the output layer by layer for $q$'s initialization .

The whole training process consists of $3$ phases. First, disable activation quantization and only train the quantization of weights. Second, fix the quantization of weights and only train the quantization of activations. Third, release quantization of both weights and activations until the model converges.

\textbf{AlexNet}: This network has five convolutional
layers and two fully connected layers. This network is the mostly used benchmark for the quantization
of neural networks. As in \cite{xnor-net,twn,admm}, we use AlexNet coupled with batch
normalization  \cite{bn} layers. We update the model by stochastic gradient descent (SGD) with the
momentum set to $0.9$. The learning rate is initialized by $0.001$ and decayed by
$0.1$ at epochs $25$ and $40$ respectively. The model is trained for at most
$55$ epochs in total.
The weight decay is set to $5e^{-4}$.
The temperature $T$ is set to $10$ and increased linearly w.r.t. the training
epochs, i.e., $T = epoch \times 10$.
Gradients are clipped with a maximum L2 norm of $5$.

The results of different quantization methods are shown in Table~\ref{tab:alexnet}. $1/1$ denotes both weights and activations are binary quantized.
As shown, our quantization network outperforms state-of-the-art methods in both weight
quantization and activation quantization.
Moreover, our quantization network is highly flexible. It is suitable for arbitrary bits quantizaion and can be applied for quantization of both weights and activation.

\textbf{ResNet}: The most common baseline architectures, including AlexNet, VGG
and GoogleNet, are all over-parameterized by design for accuracy improvements.
Therefore, it is easy to obtain sizable compression of these architectures with
a small accuracy degradation. A more meaningful benchmark would be to quantize
model architectures that are already with efficient parameters, e.g., ResNet. We
use the ResNet-18 and ResNet-50 proposed in \cite{ResNet}.

The learning rate is decayed by $0.1$ at epochs $30$ and $45$, and
the model is trained for at most $55$ epochs in total. The weight decay is set to $1e^{-4}$.
The temperature $T$ is set to $5$ and increased linearly w.r.t the training
epochs ($T = epoch \times 5$). The other settings are the same as these for AlexNet.
The results of different quantization methods are shown in Table~\ref{tab:resnet} and Table~\ref{tab:resnet50} for ResNet-18 and ResNet-50, respectively. We can see that the performance degradation of quantized
models is larger than that on AlexNet. This is reasonable because that the
parameters of the original model are more compact. It's worth noting that even
in such a compact model, our method still achieves lossless results with only $3$ bits. And as far as we know, we are the first to surpass the full-precision model on ResNet-18 with $3$ bits weight quantization.

\begin{figure*}[!tb]
	\centering
	\subfloat[Conv2-1]{
		\includegraphics[width=0.33\linewidth]{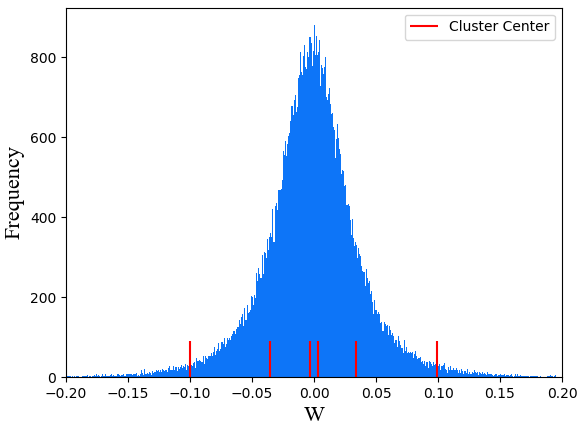}
		\label{fig:conv2}
	}
	\subfloat[Conv3-1]{
		\includegraphics[width=0.33\linewidth]{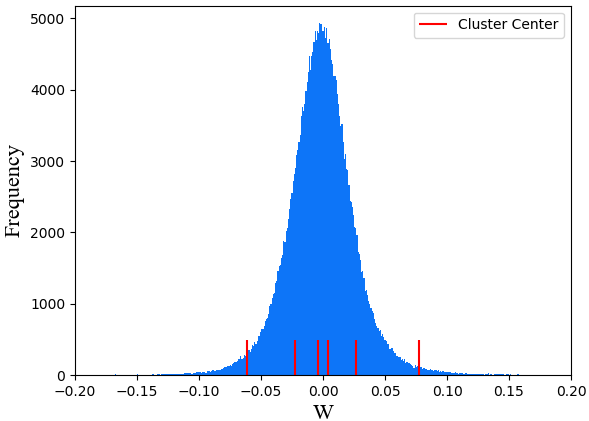}
		\label{fig:conv3}
	}
	\subfloat[Conv4-1]{
		\includegraphics[width=0.33\linewidth]{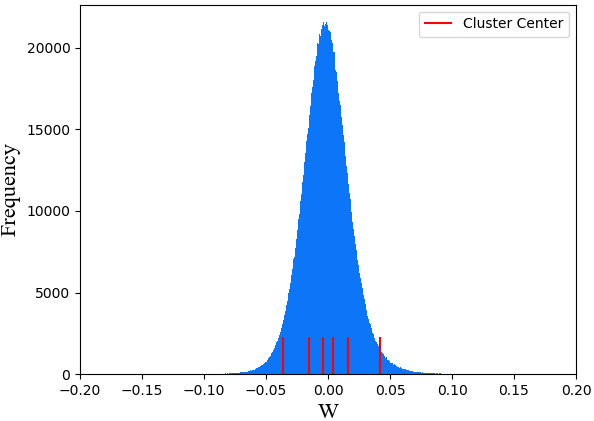}
		\label{fig:conv4}
	}
	\caption{The distribution of full-precision parameters in ResNet-18. (a)(b)(c)
		are the distributions of parameters of the first convolution layers
		from Block$2$ to Block$4$ before quantization training.}
	\label{fig:distribution}
\end{figure*}

\subsection{Object Detection}
In order to evaluate our quantization network on object detection task, we test
it on the popular architecture SSD (single shot multibox detection) \cite{ssd}. The models are trained
on Pascal VOC $2007$ and $2012$ train datasets, and are tested on Pascal VOC $2007$
test dataset. We follow the same settings in \cite{ssd} and the input images are resized
to $300\times 300$. Except the final convolutional layers with $1\times1$
kernels and the first convolution layer,
parameters of all other layers in the backbone VGG16 are quantized.

We update the model by SGD with the momentum set to $0.9$. The initial learning
rate is set to $1e^{-5}$ for quantized parameters, $1e^{-7}$ for non-quantized
parameters and decayed by $0.1$ at epochs $70$ and $90$.
Models are trained for $100$ epochs in total. The batch size is
set to $16$ and the weight decay is $5e^{-4}$. We increase the temperature $T$
by $10$ every epoch, i.e., $T = epoch \times 10$. Gradients are clipped
with maximum L2 norm of $20$.

The results are given in Table~\ref{tab:ssd}. Here we compare our model with ADMM only because other baseline quantization methods did not report their performance on object detection task.
As shown in Table~\ref{tab:ssd}, our model is slightly better than ADMM. This result is very promising since
our method is much simpler and much more general than ADMM.

\begin{table}[!tb]
	\begin{center}
		\begin{tabular}{|c|c|c|c|}
			\hline
			\diagbox{Methods}{W/A} & 2/32  & 3($\pm$4)/32 & 3($\pm$4)/8 \\
			\hline \hline
			ADMM \cite{admm} & 76.2 & 77.6 & - \\
			Ours & \textbf{76.3} & \textbf{77.7} & \textbf{76.1}  \\
			\hline
		\end{tabular}
	\end{center}
	\caption{mAP ($\%$) of SSD on Pascal VOC object detection. Performance of the
		full-precision model is $\textbf{77.8}$.}
	\label{tab:ssd}
\end{table}

\subsection{Ablation Experiments}
\label{sec:ablation-experiments}
In this section we discuss about the settings of our quantization network.
All statistics are collected from the training process of Alexnet and ResNet-18 on ImageNet.

\textbf{Configuration of Bias $b$}. Generally, the quantized values are
pre-defined linearly (e.g., $\{-1, -\frac{k-1}{k},$ $\dots, -\frac{1}{k}, 0,$
$\frac{1}{k}, \dots, \frac{k-1}{k}, 1\}$)  or logarithmically (e.g., $\{-1,
-\frac{1}{2},\dots, -\frac{1}{2^{k-1}}, 0,$ $\frac{1}{2^{k-1}}, \dots,
\frac{1}{2}, 1$) with a scale factor $\alpha$ \cite{twn,integer-net,
	admm,DoReFa,loss-aware,integer-arithmetic}.
In this paper, we find that the distribution of full-precision parameters of pre-trained model is
roughly subjected to Gaussian distribution, as shown in Fig. \ref{fig:distribution}. It indicates that
quantizing weights into linear or logarithmical intervals may not be the most suitable way.
Thus, a non-uniform quantization (e.g. K-means clustering) is adopted to counterbalance this. So, we use the $n+1$ clustering centers to determine the boundaries of quantization intervals \{$b_i$\}.
The experimental results in Table~\ref{tab:non-uniform} demonstrate the superior of non-uniform quantization over linear quantization.
We also find that adaptive learning of biases during training does not show superiority over the fixed version. Therefore,
we freeze the biases after initialization in all experiments.

\begin{table}[!tb]
	\begin{center}
		\begin{tabular}{|c|c|c|c|c|}
			\hline
			Quantization methods  & W/A  & Top-1 & Top-5 \\
			\hline \hline
			linear            & 2/32 & $60.6$ & $82.8$ \\
			non-uniform  & 2/32 & $\textbf{60.9}$ & $\textbf{83.2}$ \\
			linear           & 3($\pm$4)/32 & $60.7$ & $83.0$ \\
			non-uniform  & 3($\pm$4)/32 & $\textbf{61.9}$ & $\textbf{83.6}$ \\
			\hline
		\end{tabular}
	\end{center}
	\caption{Ablation study of training the quantization of
		AlexNet on ImageNet classification: using linear vs. non-uniform quantization.
		``W" and ``A" represent the quantization bits of weights and activations, respectively. }
	\label{tab:non-uniform}
\end{table}

\textbf{Effect of layer-wise quantization}.
As shown in Fig. \ref{fig:distribution}, the parameter magnitudes are quite different from
layer to layer (full-precision network). Therefore, it is unsuitable and less efficient to
use a shared quantization function across layers. We adopt layer-wise quantization in this paper, i.e., weights/activations from the same layer share the same quantization function and weights/activations from different layers use different quantization functions.
Table~\ref{tab:shared} shows a comparison between shared quantization function across layers and
layer-wise shared quantization function.

\begin{table}[!tb]
	\begin{center}
		\begin{tabular}{|c|c|c|c|c|}
			\hline
			Quantization methods  & W/A  & Top-1 & Top-5 \\
			\hline \hline
			shared          & 2/32 & $59.9$ & $82.4$ \\
			layer-wise     & 2/32 & $\textbf{60.9}$ & $\textbf{83.2}$ \\
			\hline
		\end{tabular}
	\end{center}
	\caption{Ablation study of training the quantization of
		AlexNet on ImageNet classification: using shared vs. layer-wise quantization
		function parameters. ``W" and ``A" represent the quantization bits of weights and activations, respectively. }
	\label{tab:shared}
\end{table}

\begin{figure}[!tb]
	\centering
	\subfloat[Top-1]{
		\includegraphics[width=0.5\linewidth]{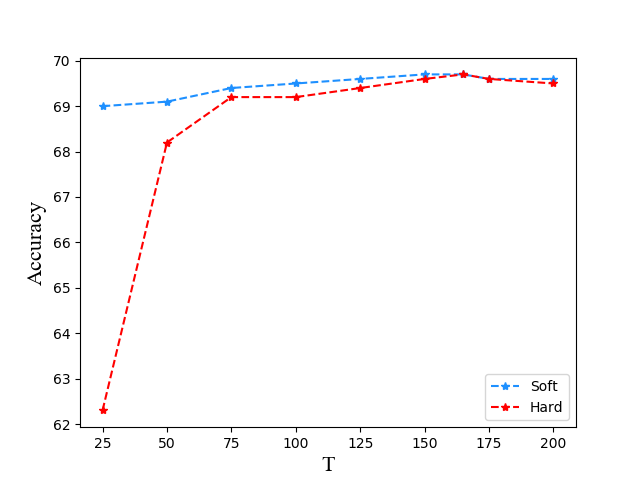}
		\label{fig:gap1}
	}
	\subfloat[Top-5]{
		\includegraphics[width=0.5\linewidth]{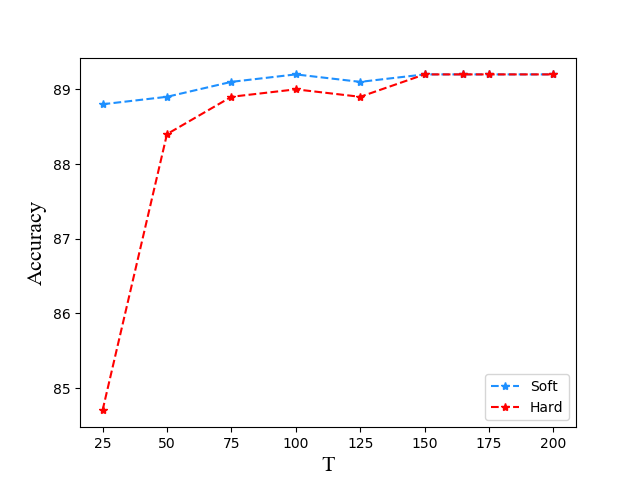}
		\label{fig:gap5}
	}
	\caption{The gap between the training model and testing model along with
		the training process for ResNet-18 $\{-4, +4\}$. The gap between
		training and testing model converges when learning proceeds.}
	\label{fig:gap}
\end{figure}

\textbf{Effect of Temperature}.
As discussed in Section \ref{sec:quantization-networks}, the temperature $T$
controls the gap between the hard quantization function Eq.
(\ref{eq:inference})
in the inference stage and the soft quantization function Eq.
(\ref{eq:quantization}) in the training stage.
In order to investigate the effect of this gap to the performance of quantized
network, we compare the testing accuracy of the models (trained with different
$T$s) when soft and hard quantization functions are adopted, as shown in Fig.
\ref{fig:gap}. We can see that as the temperature $T$ increases, the difference
between them is gradually reduced. Thus, gradually increasing temperature
$T$ during training can achieve a good balance between model learning capacity
and quantization gap.


\textbf{Training from pre-trained model}.
In our training, the temperature parameter $T$ is increased linearly w.r.t. the training epochs.
When training from scratch, the temperature $T$ may become quite large before the network is well-converged,
and the saturated neurons will slow down the network training process and make
the network stuck in bad minima.
According to Table~\ref{tab:pre-trained}, training from a pre-trained model could
greatly improve the performance compared to training from scratch.

\begin{table}[!tb]
	\begin{center}
		\begin{tabular}{|c|c|c|c|c|}
			\hline
			Training methods  & W/A  & Top-1 & Top-5 \\
			\hline \hline
			from scratch           & 3($\pm$4)/32 & $55.3$ & $78.8$ \\
			from pre-trained     & 3($\pm$4)/32 & $\textbf{70.4}$ & $\textbf{89.6}$ \\
			\hline
		\end{tabular}
	\end{center}
	\caption{Ablation study of training the quantization of
		ResNet-18 on ImageNet classification: from scratch vs. from a pre-trained
		model. ``W" and ``A" represent the quantization bits of weights and activations, respectively. }
	\label{tab:pre-trained}
\end{table}

\textbf{Time-space complexity of the final model for inference.}
Table~\ref{tab:inference-complexity} shows
the time-space complexities of the final quantization networks for inference
based on VU9P FPGA evaluation. We can see that both time and space complexity are
significantly reduced via low-bit quantization of Neural Networks.

\begin{table}[!tb]
	\begin{center}
		\begin{tabular}{|l|c|c|c|}
			\hline
			& Binary & Ternary & Full-precision \\
			\hline \hline
			Time & 1x & 1.4x & 45x \\
			Space & 1x & 2x & 32x \\
			\hline
		\end{tabular}
	\end{center}
	\caption{Time-space complexity of final inference based on VU9P FPGA
		evaluation. Each number indicates the ratio to the complexity of the binary network.
		Binary: $1$-bit weights and $1$-bit activations. Ternary: $2$-bit weights and $2$-bit activations.}
	\label{tab:inference-complexity}
\end{table}

\begin{figure}[!tb]
	\centering
	\subfloat{
		\includegraphics[width=0.33\linewidth]{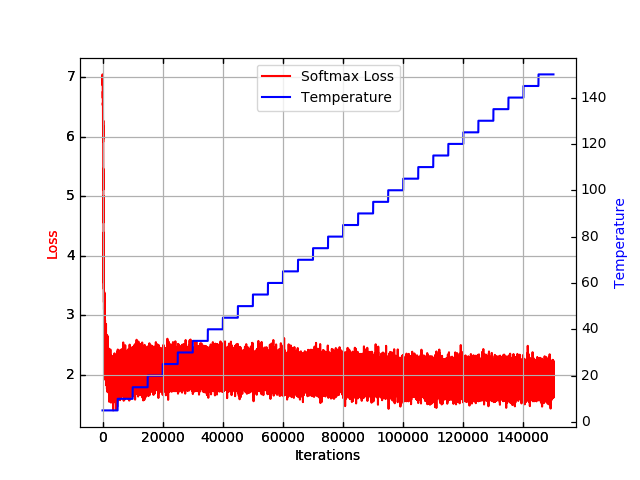}
	}
	\subfloat{
		\includegraphics[width=0.33\linewidth]{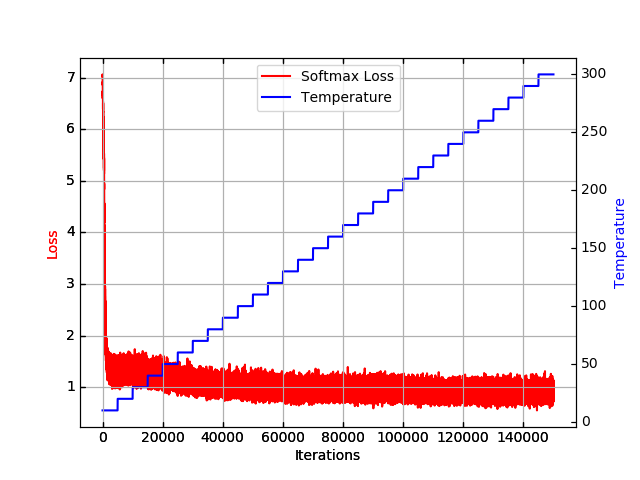}
	}
	\subfloat{
		\includegraphics[width=0.33\linewidth]{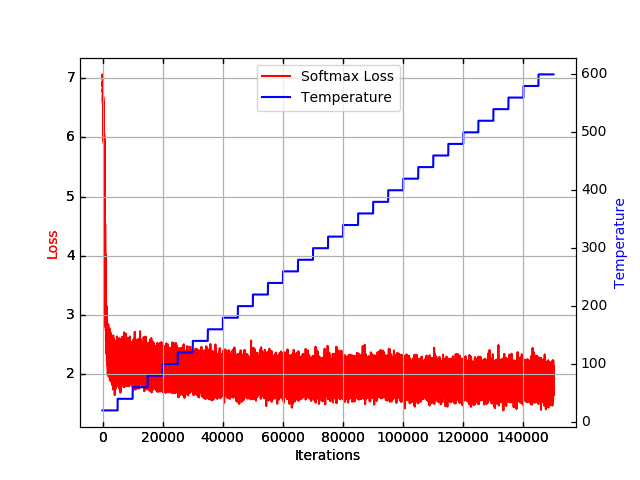}
	}

	\subfloat{
		\includegraphics[width=0.33\linewidth]{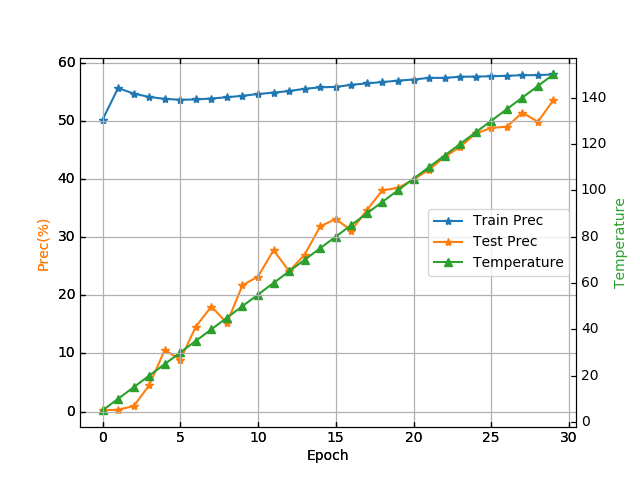}
	}
	\subfloat{
		\includegraphics[width=0.33\linewidth]{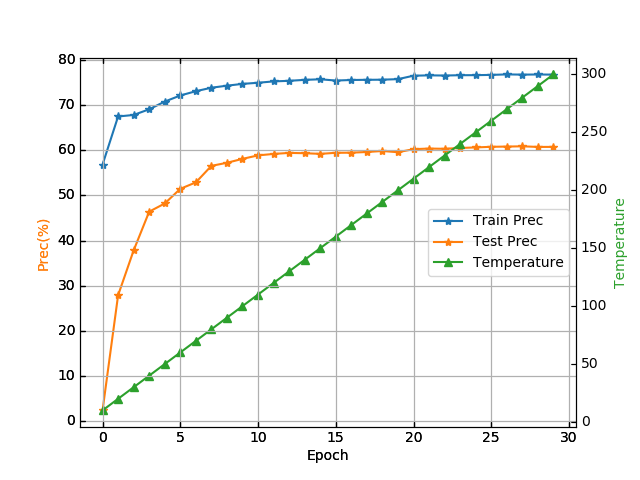}
	}
	\subfloat{
		\includegraphics[width=0.33\linewidth]{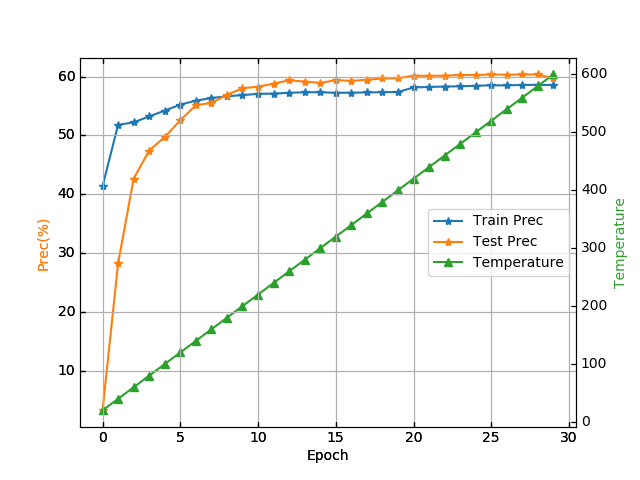}
	}
	\caption{The training error curve and the training/validation accuracy curve
		for AlexNet quantization (left to right: $T=5/10/20 * epoch$). Similar curves
		are observed for $T=1/30/40*epoch$, we do not show here because of the limit of space.}
	\label{fig:curve}
\end{figure}

\textbf{Convergence for Temprature $T$.}
The training process is very stable w.r.t. different $T$s (shown in Fig.
\ref{fig:curve}).
The approximation of the final ``soft'' quantization function to a ``hard'' step function is determined by the final temperature, which is controlled by the maximum training epoch ($T=epoch*10$).
The increasing speed of temperature (e.g.10) controls the speed of convergence (or learning rate) from a ``soft'' to ``hard'' quantization (shown in Figure $4$ in our paper), and it is consistent with the learning progress of the backbone model.
Practically, for differentbackbone models, we can tune $T$ in $\{5, 10, 20, 40\}$ via performance on validation set as the way of learning rate for DL models.

\section{Conclusion}
\label{sec:conclusions}
This work focused on interpreting and implementing low-bit quantization of deep neural networks
from the perspective of non-linear functions. Inspired by activation functions in DNNs, a soft
quantization function is proposed and incorporated in deep neural networks as a new kind of
activation function. With this differentiable non-linear quantization function embedded,
quantization networks can be learned in an end-to-end manner.
Our quantization method is highly flexible. It is suitable for arbitrary bits quantization
and can be applied for quantization of both weights and activations.
Extensive experiments on image classification and object detection tasks have verified
the effectiveness of the proposed method.

\section*{Acknowledgements}
	This work was supported in part by the National Key R\&D Program of China under contract No. 2017YFB1002203 and NSFC No. 61872329.

{\small
\bibliographystyle{ieee_fullname}
\bibliography{egbib}
}

\end{document}